\documentclass[10pt,twocolumn,letterpaper]{article}

\usepackage{iccv}
\usepackage{bm}
\usepackage{times}
\usepackage{epsfig}
\usepackage{graphicx}
\usepackage{amsmath}
\usepackage{amssymb}
\usepackage{multirow}
\usepackage[usenames,dvipsnames]{color}
\usepackage{colortbl}
\usepackage{booktabs}
\usepackage{arydshln}
\usepackage{authblk}

\usepackage[pagebackref=true,breaklinks=true,colorlinks,citecolor=blue,linkcolor=blue,bookmarks=false]{hyperref}
\iccvfinalcopy % *** Uncomment this line for the final submission

\usepackage{pifont} % for cmark and xmark
\newcommand{\cmark}{{\color{black}\ding{51}}}%

\newcommand{\rbf}[1]{{\color{red}\textbf{{#1}}}}
\newcommand{\bif}[1]{{\color{blue}\textit{{#1}}}}
\definecolor{mygray}{gray}{.9}

\def\iccvPaperID{9923} % *** Enter the ICCV Paper ID here

\graphicspath{{figures/}}
\ificcvfinal\fi

\begin{document}

%%%%%%%%% TITLE
\title{ 
The Road to Know-Where: An Object-and-Room Informed Sequential BERT for Indoor Vision-Language Navigation}

\author{\vspace{-5mm}
	{\fontsize{10.5}{10.5}\selectfont Yuankai Qi$^{1}$\quad Zizheng Pan$^2$\quad Yicong Hong$^3$\quad Ming-Hsuan Yang$^{4,5,6}$\quad Anton van den Hengel$^1$\quad Qi Wu$^1${\thanks{Corresponding author}}}\\
\vspace{-3mm}
{\fontsize{10.8}{10.8}\selectfont $^1$Australian Institute for Machine Learning, The University of Adelaide\,\,\, $^2$Monash University}\\
{\fontsize{10.8}{10.8}\selectfont $^3$The Australian National University\,\,\,  $^4$University of California, Merced \,\,\, $^5$Google Research \,\,\,$^6$Yonsei University}\\
{\tt\small \{qykshr, zizhpan\}@gmail.com \quad yicong.hong@anu.edu.au \quad mhyang@ucmerced.edu \quad  \{anton.vandenhengel, qi.wu01\}@adelaide.edu.au}\\
\vspace{-3mm}
}

\maketitle
\ificcvfinal\thispagestyle{empty}\fi

%%%%%%%%% ABSTRACT
\begin{abstract}
Vision-and-Language Navigation (VLN) requires an agent to find a path to a remote location on the basis of natural-language instructions and a set of photo-realistic panoramas. Most existing methods take the words in the instructions and the discrete views of each panorama as the minimal unit of encoding. However, this requires a model to match different nouns (e.g., TV, table) against the same input view feature. In this work, we propose an object-informed sequential BERT to encode visual perceptions and linguistic instructions at the same fine-grained level, namely objects and words. 
Our sequential BERT also enables the visual-textual clues to be interpreted in light of the temporal context, which is crucial to multi-round VLN tasks. Additionally, we enable the model to identify the relative direction (e.g., left/right/front/back) of each navigable location and the room type (e.g., bedroom, kitchen) of its current and final navigation goal, as such information is widely mentioned in instructions implying the desired next and final locations. We thus enable the model to know-where the objects lie in the images, and to know-where they stand in the scene.
Extensive experiments demonstrate the effectiveness compared against several state-of-the-art methods on three indoor VLN tasks: REVERIE, NDH, and R2R. 
Project repository: \href{https://github.com/YuankaiQi/ORIST}{https://github.com/YuankaiQi/ORIST}
\end{abstract}

%%%%%%%%% BODY TEXT
\section{Introduction}
Vision-and-Language Navigation (VLN) offers the appealing prospect of more flexible interactions with robotic applications including domestic robots and personal assistants. 
One of the first VLN tasks to appear was Room-to-Room navigation~(R2R)~\cite{r2r}.  This task saw an agent initialised at a random location within a simulated environment rendered from real images, and required it to navigate to a remote goal location according to natural-language instructions, such as ``Leave the bedroom, and enter the kitchen. Walk forward, and take a left at the couch. Stop in front of the window.'' 
The actions available to the agent at each step are to investigate the current panorama, to move to a neighbouring navigable location/viewpoint, or to stop.
An interactive version of the problem is introduced in~\cite{hanna,ndh}, while REVERIE~\cite{reverie} extends it to identifying remote objects, and TOUCHDOWN~\cite{touchdown} introduces outdoor environments.

\begin{figure}[!t]
\begin{center}
	\includegraphics[width=\linewidth]{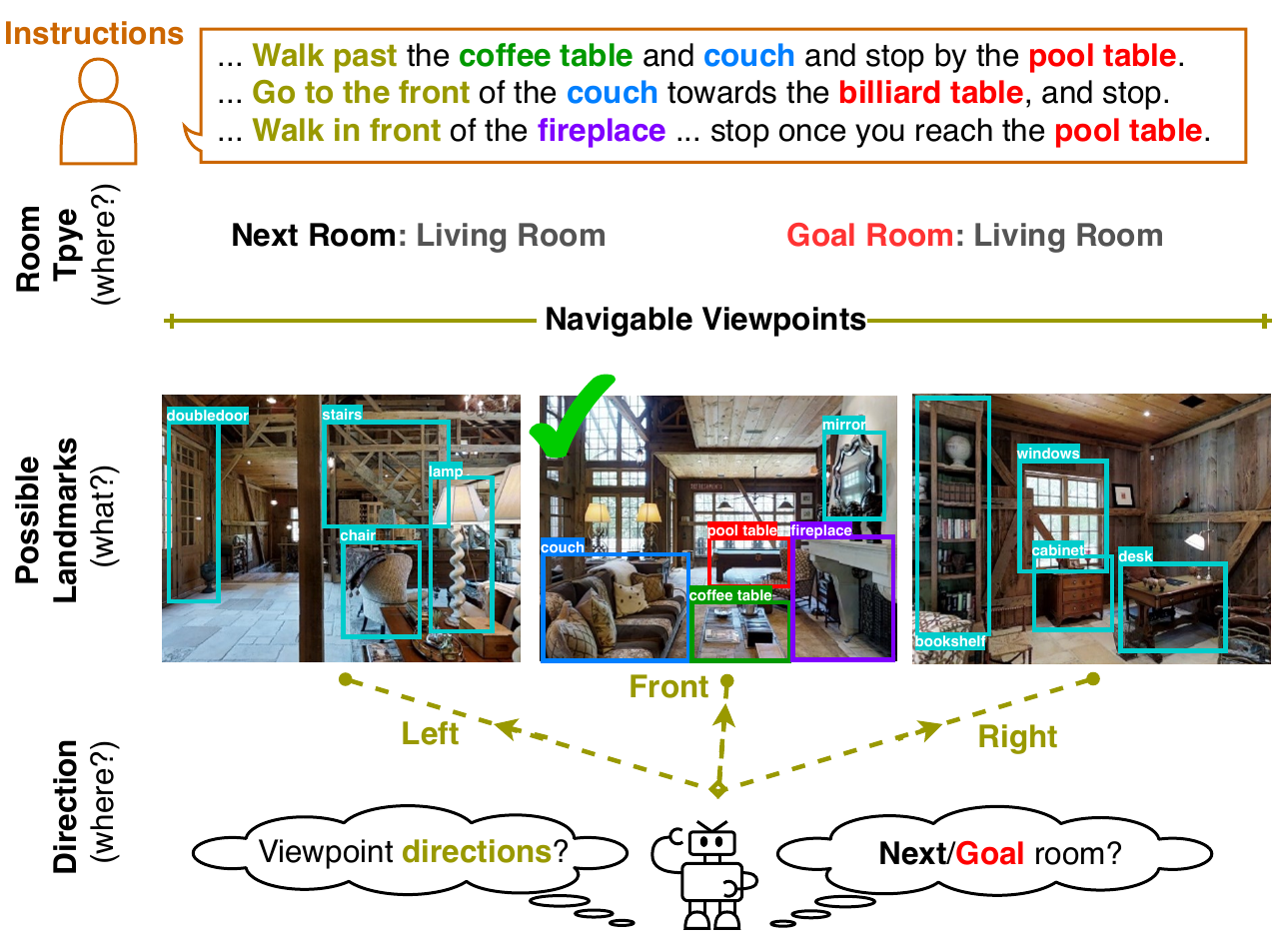}
\end{center}
\caption
{Objects, rooms, and directions are important clues that can be inferred from visiolinguistic information. Our {\bf{O}}bject-and-{\bf{R}}oom {\bf{I}}nformed {\bf{S}}equential BER{\bf{T}} (ORIST) is designed to learn to navigate by leveraging this information.
}
\label{fig:firstFig}
\end{figure}

Numerous methods have been proposed to address indoor VLN tasks.
Ma \etal~\cite{selfmonitor} propose to learn textual-visual co-grounding to enhance the understanding of instructions completed in the past and the instruction to be executed next.
Heuristic search algorithms for exploration and back-tracking are introduced in~\cite{fast,regretful}. 
Qi \etal~\cite{oaam} propose to disentangle object- and action-related instruction chunks for more accurate visual-textual matching.
Another line of work exploits data augmentation to improve the generalisation ability in unseen environments. 
In~\cite{speakerfollower}, a speaker model is proposed to generate instructions for newly sampled trajectories, and in~\cite{envdrop} visual information of seen scenarios is randomly dropped to mimic unseen scenarios. 
Most recently, it has been empirically demonstrated that hard example sampling and auxiliary losses~\cite{advsampling,auxiliary} are helpful for navigating unseen environments.

Although significant advances have been made, these methods take words in instructions and each discrete view of a panorama as the minimal unit for encoding, which limits the ability to learn the relationships between language elements and fine-grained visual entities. 
This is because the view feature, generally obtained  by ResNet~\cite{resnet} which is trained for whole-image classification, mainly represents one salient object but it needs to be matched to many different natural-language landmarks mentioned in crowd-sourced navigation instructions. 
Figure \ref{fig:firstFig} shows an example where the \textit{coffee table}, \textit{couch}, and \textit{fireplace} are used as navigation landmarks in different instructions, but they need to match to the same input view feature in most existing VLN methods. 

Motivated by the success of BERT-like models (Bidirectional Encoder Representations from Transformers)~\cite{uniter,vilbert,vl-bert} in joint textual and visual understanding, 
we propose an object-and-room informed sequential BERT to encode instructions and visual perceptions at the same fine-grained level, namely words and object regions. 
This enables the model to better ``know where'' the objects referred to lie.
We also introduce temporal context into our BERT-based model, which allows the model to be aware of completed parts of instructions and seen environments, resulting in more accurate next action prediction.

Relative directions (\ie, left, right, front, and back) and room types (\eg, living room, bedroom) are important clues for VLN tasks as they provide strong directional guidance and semantic-visual information for action selection. 
To take advantage of such information we incorporate a direction loss and two room-type losses into the proposed model.  These predict the relative direction of navigable viewpoints, the type of room that need to be reached next, as well as the type of room at the final goal location. 
The relative direction is predictable from each viewpoint's orientation description (\ie, heading and elevation angles),
and the room types are identifiable through the presence of specific objects, such as a couch indicating living room or a microwave indicating kitchen.
This provides the model with opportunity to ``know where'' it is, and where it is going.

To demonstrate the generalisation ability of our model, we evaluate  on three different indoor VLN tasks: remote object grounding (REVERIE) \cite{reverie}, visual dialog navigation (NDH) \cite{ndh} and room-to-room navigation (R2R) \cite{r2r}. 
Our method achieves state-of-the-art results on all of the listed tasks: 18.97 SPL (9.28 RGSPL) on the REVERIE task, 3.17 GP on the NDH task, and 52$\%$ SPL on the R2R task.

\begin{figure*}[!t]
\centering
\includegraphics[width=\linewidth]{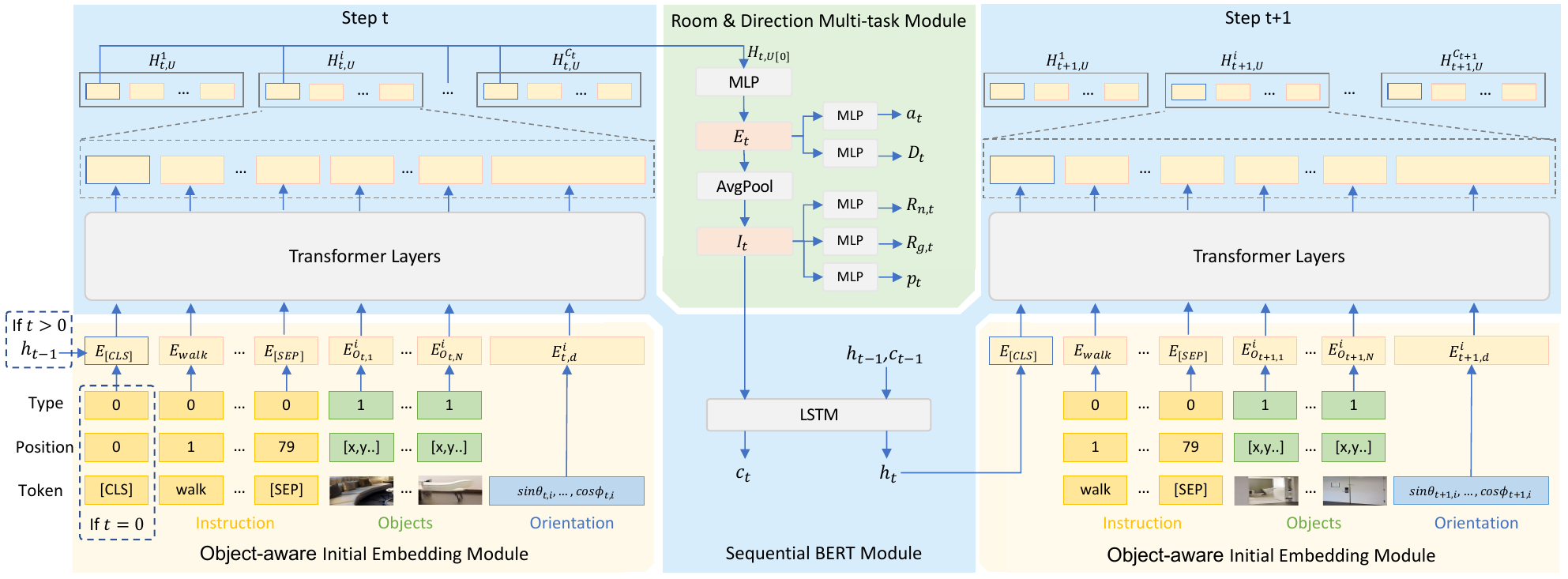}
\caption{An unrolled illustration of the proposed {\bf{O}}bject-and-{\bf{R}}oom {\bf{I}}nformed {\bf{S}}equential BER{\bf{T}} (ORIST), which is composed of three main parts: the object-level initial embedding module (Section \ref{sec:initalEmb}), the sequential BERT module (Section \ref{sec:modelStem}), and the room-and-direction multi-task module (Section \ref{sec:room-awareLoss}). Note that all modules are reused between steps.
}
\label{fig:model_arch}
\vspace{-3mm}
\end{figure*}

\section{Related Work}
In this section, we briefly review closely related VLN methods and vision-and-language BERT-based works. 
\vspace{-6mm}
\paragraph{Vision-and-Language Navigation.}
A large number of existing works have focused on cross-modality matching and generalisation from seen to unseen environments.
For cross-modality matching, Ma~\etal~\cite{selfmonitor} propose a visual-textual co-grounding module and a progress monitor to distinguish completed instructions from those yet to be executed. In \cite{rcm}, a local navigation distance reward and a global instruction-trajectory matching reward are employed for reinforcement learning. Qi~\etal~\cite{oaam} propose a multi-module framework to separately learn matching between action-related words and candidate orientations, and between landmark-related words and candidate visual features.
In NvEM~\cite{nvem}, visual features of navigable locations are adaptively enhanced by neighbor views leading to better visual-textual matching.
To improve generalisation, data augmentation is introduced in \cite{speakerfollower} and  \cite{envdrop} to generate new data from seen environments. Fu \etal propose to mine hard training samples via adversarial learning in~\cite{advsampling}. In~\cite{softexp}, reinforcement learning based on adaptively learned rewards is utilised to enhance generalisation. On the other hand, active exploration is designed in~\cite{avig} to learn when, where, and what information to explore so as to form a robust agent.

In contrast to these methods, which take words and discrete images  within each panorama as the minimal unit for encoding,
our model is designed to handle fine-grained inputs, namely objects and words, to facilitate coordinated textual and visual understanding.

\vspace{-6mm}
\paragraph{Vision-Language BERTs.} VL-BERTs have been widely adopted to learn joint visual and textual 
models~\cite{vl-bert,vilbert,lxmert,uniter,oscar} for Vision-and-Language tasks, such as Visual Question Answering~\cite{vqa}, Visual Commonsense Reasoning~\cite{vcr} and Referring Expressions~\cite{refer_exp}. In VLN, Majumdar
\etal~\cite{bertvln} predict whether a trajectory matches an instruction using a transformer. However, this method does not predict navigation actions. On the other hand, as pointed out in~\cite{prevalent}, most existing VL-BERTs learn to match textual and visual elements without considering their relationship to the actions available. 
To solve the problem, Hao \etal~\cite{prevalent} propose a model that is pre-trained with image-text-action triplets generated from the R2R dataset. However, the pre-trained model only acts as an instruction encoder and therefore represents a pre-processing step for other VLN methods.
R-VLNBERT~\cite{rvlnbert} enhances existing VLBERTs for VLN by representing agent states with the built-in [CLS] token, while it learns word-image level relationship.

One of the key distinguishing features of the model proposed here is that, unlike most of existing VL-BERTs that are designed for one-round decision tasks (\eg, VQA), 
our model is endowed with temporal context and thus is suitable for partially observed Markov decision processes. 

\section{Proposed Method}

In VLN tasks, a robot agent is given a natural language instruction $\mathcal{X}=\{x_1, x_2, \cdots, x_L\}$, where $L$ is the length of the instruction and $x_i$ is a single word token. 
At each step, the agent is provided with a panoramic RGB image, which is divided into 36 discrete views $\mathcal{V}=\{v_{t,i}\}_{i=1}^{36}$ (12 horizontal by 3 vertical). 
A navigable location falls in one of these views and it has an orientation $\mathcal{D}_{t,i}=\langle\sin{\theta_{t,i}}, \cos{\theta_{t,i}}, \sin{\phi_{t,i}}, \cos{\phi_{t,i}}\rangle$, where $\theta$ and $\phi$ are measured relative to the current heading and elevation. %}
At each step, the agent predicts the next navigation action by selecting from a candidate set of navigable locations (including the current location).
The agent stops if the current location is selected or it reaches the maximum number of steps.

\subsection{Overview}

Figure \ref{fig:model_arch} shows an unrolled illustration of the proposed {\bf{O}}bject-and-{\bf{R}}oom {\bf{I}}nformed {\bf{S}}equential  BER{\bf{T}} (ORIST).
The ORIST is mainly composed of three parts: an object-level initial embedding module, a sequential  BERT module and a room-and-direction multi-task module. All these modules are reused between steps.
At step $t$, the initial embedding module takes as inputs an instruction $\mathcal{X}$, features of object regions $\mathcal{O}_{t}$, and navigable orientations $\mathcal{D}_{t}$.  
Then its outputs are fed into transformer layers of the sequential  BERT module to adaptively fuse information from individual words and visual objects as well as each orientation via the self-attention mechanism. 
Finally, the fused features $\mathbf{H}_{t,U}$ are sent to two branches. One branch is the room-and-direction multi-task module, which predicts the room type $\mathbf{R}_{n,t}$ and $\mathbf{R}_{g,t}$, the direction $\mathbf{D}_t$, the next navigating action $\mathbf{a}_t$, and the navigation progress $\mathbf{p}_t$. Another branch is the LSTM of the sequtial BERT module, which produces a new temporal context $\mathbf{h}_t$ and connects to the next navigation step. The whole model is trained end-to-end.

\subsection{Object-level Initial Embedding Module}
\label{sec:initalEmb}

Existing VLN methods typically use ResNet~\cite{resnet} features of each discrete view
to represent visual information. 
However, this may weaken vision-and-language matching because the ResNet is trained for image classification which may represent only one salient object, leading to the same image feature being matched to a variety of different objects mentioned in instructions (\eg, laptop and table).  
To address this issue, we introduce an object-based representation  to facilitate object-level cross-modality matching. 
The bounding boxes are adopted from REVERIE~\cite{reverie}.

When designing the embedding module, we take into account the following three considerations.
(I) The distinct characteristics of natural-language instructions and visual objects, \eg,
word tokens \vs object regions, word order \vs object position.  
(II) The type of an input token (\eg, a visual or textual input).
(III) The characteristic of VLN tasks, \eg, the orientation feature of each navigable location. 
To this end, for an instruction $\mathcal{X}$ and a candidate location in view $v_{t,i}$ composed of $N^i$ objects $\mathcal{O}_{t,i}=\{o_{t,i}^1,\cdots,o_{t,i}^{N^i}\}$, we obtain the initial embedding via
\begin{small}
\begin{align}
\mathbf{E}_{t,w} & = \mathrm{Emb}(\bm{x}^{tok}) + \mathrm{Emb}(\bm{x}^{pos}) + \mathrm{Emb}(\bm{x}^{type}), \nonumber \\
\mathbf{E}_{t,o}^i & = \mathrm{FC}(\bm{o}_{t,i}^{fea}) + \mathrm{FC}(\bm{o}_{t,i}^{pos}) + \mathrm{Emb}(\bm{o}_{t,i}^{type}), \nonumber \\
\mathbf{E}_{t,d}^i & = \mathrm{FC}(\bm{d}_{t,i}),
\label{eq:embedding}
\end{align}
\end{small}%
where $\mathbf{E}_{t,w}\in\mathbb{R}^{(L+2)\times d_h}$ denotes the embedding of the expanded instruction $\bm{x}^{tok}=\{[\mathrm{CLS}], \mathcal{X}, [\mathrm{SEP}]\}$. 
In addition, 
$\bm{x}^{pos}=[0,1,\cdots]$ is zero-based token position;
$\mathbf{E}_{t,o}^i\in\mathbb{R}^{N^i\times d_h}$ denotes an embedding of $N^i$ object regions;
$\bm{o}_{t,i}^{fea}\in\mathbb{R}^{N^i\times2048}$ represents objects'  features extracted using FasterRCNN as described in~\cite{bottomup}; %
and 
$\bm{o}_{t,i}^{pos}\in\mathbb{R}^{N^i\times7}$ is the position feature of $N$ object regions, of which each row is composed of the $x, y$ location of the top-left and bottom-right points, and the region's height, width, and area.
In Eq. \eqref{eq:embedding},
$\bm{x}^{type}=\mathbf{0}$ and  $\bm{o}_{t,i}^{type}=\mathbf{1}$ are  token type vectors, and
$\mathbf{E}_{t,d}^i\in\mathbb{R}^{1\times d_h}$ denotes the   embedding  of the candidate location's orientation 
$\bm{d}_{t,i}\in\mathbb{R}^{1\times4}=\mathcal{D}_{t,i}$. 
$\mathrm{Emb}(\cdot)$ is a projection layer that embeds tokens to feature vectors. $\mathrm{FC}(\cdot)$ is a fully-connected layer.

For each  candidate location $i$, we obtain its initial embedding matrix $\mathbf{H}_{t,0}^{i}=[\mathbf{E}_{t,w}, \mathbf{E}_{t,o}^i, \mathbf{E}_{t,d}^i]\in\mathbb{R}^{C_t\times d_h}$, where $C_t$ is the maximum number of tokens at step $t$ across all candidates.
The subscript $0$ in $\mathbf{H}_{t,0}^i$ denotes it is an initialisation before going through the sequential BERT module (detailed in the next section).
The concatenation of all $G_t$ candidates at step $t$ constructs the initial embedding   $\mathbf{H}_{t,0}=[\mathbf{H}_{t,0}^1;\cdots;\mathbf{H}_{t,0}^{G_t}]\in\mathbb{R}^{G_t \times C_t \times d_h}$.
\subsection{Sequential  BERT Module} 
\label{sec:modelStem}
BERT-like models have demonstrated their effectiveness in learning textual-visual entity matching~\cite{uniter,vl-bert}.
However, most of them are designed for one-round decision tasks (\eg, VQA). 
VLN, as a multi-round task and partially observed Markov decision process, requires a model to be aware of which parts of an instruction have been completed and which parts have yet to be executed.
As such, a direct application of BERTs to VLN does not perform well due to the loss of temporal context. 
To address these problems, in this work we propose a sequential BERT module.

The light blue panel in Figure~\ref{fig:model_arch} shows an unrolled illustration of our sequential BERT at step $t$ and $t+1$, which consists of two main components: transformer layers and an LSTM. Each transformer layer $\mathcal{F}_j$ is a basic transformer~\cite{transformer}, which  is   a multi-head self-attention  
\begin{equation}
\mathbf{H}_{t,j}=\mathcal{F}_j(\mathbf{H}_{t,j-1},\mathbf{M}_t),
\end{equation}
where $\mathbf{H}_{t,j-1}\in\mathbb{R}^{G_t\times C_t\times d_h}$ is the output from a previous transformer layer, $\mathbf{M}_t$ is a mask matrix indicating whether a token can be attended, and $\mathbf{H}_{t,j}=[\mathbf{A}_{t,j,1}, \cdots, \mathbf{A}_{t,j,n}]$ is a concatenation of outputs of $n$ attention heads. 
Concretely, the $k$-th head's output $\mathbf{A}_{t,j,k}$  is obtained by
\begin{small}
\begin{gather}
\mathbf{A}_{t,j,k} = \textrm{softmax}(\frac{\mathbf{Q}^\top\mathbf{K}}{\sqrt{d_{j,k}}}+\mathbf{M}_t)\mathbf{V}^\top,\nonumber\\
\mathbf{Q}=\mathbf{W}_j^Q\mathbf{H}_{t,j-1}^\top, \mathbf{K}=\mathbf{W}_j^K\mathbf{H}_{t,j-1}^\top, \mathbf{V}=\mathbf{W}_j^V\mathbf{H}_{t,j-1}^\top,
\end{gather}
\end{small}% 
where $\mathbf{W}_j^*$ is a learnable parameter matrix, and $d_{j,k}$ is the embedding dimension of  $\mathbf{A}_{t,j,k}$.
Assume we have $U$ transformer layers in total, as was demonstrated experimentally in~\cite{reveal,uniter,vilbert}, then the embedding of the first token (\ie, [CLS]) of the last transformer layer, denoted as $\mathbf{H}_{t,U}[0]\in\mathbb{R}^{G_t\times d_h}$, is able to represent fused mutually attended information from textual and visual inputs.

To enable these transformer layers to model temporal context, an intuitive solution is to use a collection of object entities observed along the navigation path.
In this way, one can expect that previously observed objects might be matched to completed instruction parts, and  new visual perceptions could be matched to instruction parts to be executed. However, this will involve a huge computation cost (\eg, GPU memory and computation time) as the agent navigates.
To tackle this problem, we propose to employ an LSTM to adaptively learn how to summarise seen instructions and visual perceptions as the agent navigates. 
Specifically, we first encode $\mathbf{H}_{t,U}[0]$, which contains mutually attended instruction and visual information, to get a deeper representation $\mathbf{E}_t\in\mathbb{R}^{G_t\times d_h}$ via
\begin{equation}
\mathbf{E}_t=\textrm{Tanh}(\mathbf{W}_{e}\mathbf{H}_{t,U}[0]).
\end{equation}
Then,  we aggregate surrounding information from all the $G_t$ candidates via 
$
\mathbf{I}_t=\textrm{AvgPool}(\mathbf{E}_{t}).
$
Next, we obtain the new temporal context $\mathbf{h}_t$ by
\begin{equation}
\mathbf{h}_t, \mathbf{c}_t=\textrm{LSTM}(\mathbf{I}_{t},\mathbf{h}_{t-1}, \mathbf{c}_{t-1}), 
\end{equation}
where $\mathbf{h}_{t-1}$ and $\mathbf{c}_{t-1}$ represents previous temporal context and LSTM cell state, respectively.  $\mathbf{h}_{0}$ and $\mathbf{c}_{0}$ are initialised with zeros. 
Thus, $\mathbf{h}_t$  contains the accumulated temporal context along the navigation and it is passed to the next navigation step as shown by the right panel of Figure~\ref{fig:model_arch}.

\subsection{Room-and-Direction Multi-task Module}
\label{sec:room-awareLoss}
Room types and directions can be useful information for navigation as they are often mentioned in instructions and provide strong navigation guidance.
We note that a certain room type can usually be characterised by the presence of specific objects, such as bed for bedroom, TV/couch for living room, oven/microwave for kitchen. 
As our model takes objects as inputs, it is thus able to predict which room the agent should go to according to the current visual perception and instruction. 
Furthermore, it is also important for agents to understand the room that constitutes its final goal. 
This can help the agent make decisions based on the long-term goal.
Additionally, we enable an agent to recognise the relative direction (\ie, left/right/front/back) of each candidate in order to better align with key direction cues (\eg, ``turn left/right'') in instructions.
The progress monitor~\cite{auxiliary} is also adopted to further facilitate analysis of progress.
We use $\mathrm{MLP}$ layers to implement the above goals:
\begin{gather}
\mathbf{D}_t=\mathrm{MLP}(\mathbf{E}_t), \nonumber \\
\mathbf{R}_{n,t}=\mathrm{MLP}(\mathbf{I}_t),\; \mathbf{R}_{g,t}=\mathrm{MLP}(\mathbf{I}_t), \nonumber \\
\mathbf{a}_t=\mathrm{MLP}(\mathbf{E}_t), \;
\mathbf{p}_t=\mathrm{MLP}(\mathbf{I}_t),
\label{eq:MLP}
\end{gather}
where $\mathbf{D}_t, \mathbf{R}_{n,t}, \mathbf{R}_{g,t}, \mathbf{a}_t, \mathbf{p}_t$ denote the relative direction prediction for each candidate (we evenly divide the surrounding 360$^{\circ}$ into four directions representing left/right/front/back), the room type prediction for the next step and the goal location,  the action logit, and the navigation progress, respectively. The ground truth room types are extracted from the Matterport3D dataset~\cite{mattport}.
$\mathbf{E}_t$ is composed of embeddings of all candidate  and $\mathbf{I}_t$ is the fused embedding of all candidates as described in Section~\ref{sec:modelStem}.

\paragraph{Loss Function.}
Similar to \cite{envdrop}, we adopt the training strategy of mixed Imitation Learning (IL) and Reinforcement Learning (RL).
Based on the above-mentioned tasks, for imitation learning we have the following losses:
\begin{equation}
\mathcal{L}^{IL} =  \mathcal{L}^{\mathbf{D}} +  \lambda_1 \mathcal{L}^{\mathbf{R}_n} + \lambda_2 \mathcal{L}^{\mathbf{R}_g} + \mathcal{L}^{\mathbf{a}} + \mathcal{L}^{\mathbf{p}}, 
\label{eq:loss}
\end{equation}
where $\mathcal{L}^{\mathbf{D}}$, $\mathcal{L}^{\mathbf{R}_n}$, $\mathcal{L}^{\mathbf{R}_g}$, $\mathcal{L}^{\mathbf{a}}$ are cross-entropy losses for the relative direction prediction $\mathbf{D}_t$, the room type  of the next step $\mathbf{R}_{n,t}$, the room type  of the goal location $\mathbf{R}_{g,t}$, and the navigating action $\mathbf{a}_t$, respectively. 
$\lambda_1$ and $\lambda_2$ are trade-off factors.
$\mathcal{L}^{\mathbf{p}}$ is the progress loss adopted from \cite{auxiliary}, which is a BCELoss  $\mathcal{L}^{\mathbf{p}} = \sum_t -p_t^*\textrm{log}(\mathbf{p}_t)$ where $p_t^*$ is the teacher progress at each time step $t$. As in~\cite{envdrop}, the A2C reinforcement learning method is adopted with the loss function 
\begin{equation}
\mathcal{L}^{RL} = -\sum_t {a}_t^*\textrm{log}(p_{a_t^*})Z_t, 
\end{equation}
where $a_t^*$ is a sampled navigation action, $p_{a_t^*}$ is its probability and $Z_t=\mathrm{FC}(\mathbf{I}_t)$ is an estimated reward. 
Finally, the total loss is
\begin{equation}
\mathcal{L} = \mathcal{L}^{RL} + \lambda_3\mathcal{L}^{IL}, 
\end{equation}
where $\lambda_3$ manages the trade-off between RL and IL.

\section{Experiments and Analysis}
\label{sec:expriments}
In this section, we present extensive experimental evaluations on three VLN tasks: REVERIE~\cite{reverie}, NDH~\cite{ndh}, and R2R~\cite{r2r}. 
We first give a brief introduction to these tasks and the evaluation protocols, and then present two sets of experimental results: one is  a comparison against several state-of-the-art VLN methods, and the other one is an ablation study on the proposed model.

\subsection{Evaluation Tasks}

% \vspace{-2pt}
\paragraph{REVERIE.} This task~\cite{reverie} requires an agent to localise a remote target object within a photo-realistic indoor environment on the basis of concise human instructions, such as ``Go to the massage room with the bamboo plant and cover the diamond shaped window''. REVERIE contains two sub-tasks: (I) Vision-Language Navigation, where the agent needs to navigate to the target room; (II) Referring Expression Grounding, where the agent must identify the target object of an interaction from a provided set of objects (the interaction is not required).  
Here we compare primarily against the navigation sub-task, as our model has been developed for navigation rather than grounding.

\paragraph{NDH.} The Navigation from Dialog History (NDH) task is derived from the Cooperative Vision-and-Dialogue Navigation (CVDN) dataset~\cite{ndh}.
Specifically, each item of the CVDN dataset is a trajectory paired with a navigation dialogue between an {Oracle} and a {Navigator}, where the Oracle provides additional instruction towards navigating to the goal location when Navigator requests help. Each round of dialogue (together with previous dialogue if it exists) in CVDN is extracted, and forms an item within the NDH dataset.  Based on which path is selected as the ground-truth, NDH has three settings: (I) the {Oracle} setting, which uses the shortest path observed by the Oracle; (II) the {Navigator} setting, which uses the path taken by Navigator; and (III) the {Mixed} setting, which takes the path of Navigator if it visits the target location, or the shortest path if not.

\paragraph{R2R.} This task~\cite{r2r} requires an agent to follow natural language instructions to navigate from one room to a remote goal room in a photo-realistic indoor environment. Instructions in R2R contain rich linguistic information about the navigation trajectory, such as ``Walk through the kitchen. Go past the sink and stove, stand in front of the dining table on the bench side.''

% ---------- REVERIE -----------
\begin{table*}[!t]
\centering
\resizebox{\linewidth}{!}{
	\begin{tabular}{l|cccccc||cccccc||cccccc}
		\toprule
		\multicolumn{1}{c}{\multirow{3}{*}{Methods}}  & \multicolumn{6}{|c||}{Val Seen} &\multicolumn{6}{c||}{Val UnSeen} & \multicolumn{6}{c}{Test}\\ \cline{2-19}& \multicolumn{4}{c|}{Navigation}  & \multicolumn{1}{c}{\multirow{2}{*}{RGS$\uparrow$}}&\multicolumn{1}{c||}{\multirow{2}{*}{\textbf{RGSPL}$\uparrow$}} & \multicolumn{4}{c|}{Navigation}  & \multicolumn{1}{c}{\multirow{2}{*}{RGS$\uparrow$}}&\multicolumn{1}{c||}{\multirow{2}{*}{\textbf{RGSPL}$\uparrow$}} & \multicolumn{4}{c|}{Navigation}   & \multicolumn{1}{c}{\multirow{2}{*}{RGS$\uparrow$}}&\multicolumn{1}{c}{\multirow{2}{*}{\textbf{RGSPL}$\uparrow$}} \\
		\cline{2-5} \cline{8-11} \cline{14-17}  & SR$\uparrow$ & OSR$\uparrow$   & \textbf{SPL}$\uparrow$  & \multicolumn{1}{c|}{TL$\downarrow$} &  &  & SR$\uparrow$ & OSR$\uparrow$ & \textbf{SPL}$\uparrow$ &\multicolumn{1}{c|}{TL$\downarrow$} &  & & SR$\uparrow$ & OSR$\uparrow$ & \textbf{SPL}$\uparrow$ & \multicolumn{1}{c|}{TL$\downarrow$} &    & \\ 
		\hline
		RCM~\cite{rcm} &23.33& 29.44 & 21.82& \multicolumn{1}{c|}{10.70} & 16.23& 15.36  & 9.29 & 14.23& 6.97 &\multicolumn{1}{c|}{11.98}     & 4.89& 3.89  & 7.84  &11.68 & 6.67 & \multicolumn{1}{c|}{10.60} & 3.67&  3.14\\
		
		SM~\cite{selfmonitor} & 41.25& 43.29  & 39.61& \multicolumn{1}{c|}{7.54}   & 30.07 &  \bif{28.98} & 8.15 & 11.28 &6.44 & \multicolumn{1}{c|}{9.07} & 4.54& 3.61 & 5.80& 8.39 & 4.53 &\multicolumn{1}{c|}{9.23}   & 3.10& 2.39 \\
		
		FAST-Short~\cite{fast} & 45.12& 49.68 &40.18& \multicolumn{1}{c|}{13.22}  &31.41 & 28.11 & 10.08 & 20.48 & 6.17 & \multicolumn{1}{c|}{29.70}  & 6.24 & 3.97 & 14.18 & 23.36 & 8.74 & \multicolumn{1}{c|}{30.69}  & 7.07 & 4.52 \\ 
		
		Nav-Pointer~\cite{reverie} & 50.53 & 55.17  & \rbf{45.50} & \multicolumn{1}{c|}{{16.35}}  & 31.97 & \rbf{29.66} & 14.40 & 28.20  & \bif{7.19} & \multicolumn{1}{c|}{{45.28}}  & 7.84 & \bif{4.67} & 19.88 & 30.63 & \bif{11.61} & \multicolumn{1}{c|}{{39.05}}  & 11.28 & \bif{6.08} \\
		\hline
		
		ORIST & 45.19  & 49.12   & \bif{42.21}  & \multicolumn{1}{c|}{10.73 }& 29.87  & 27.77  & 16.84  & 25.02  & \rbf{15.14}  & \multicolumn{1}{c|}{10.90}     & 8.52& \rbf{7.58}  & 22.19 & 29.20  & \rbf{18.97}  & \multicolumn{1}{c|}{11.38} & 10.68 &  \rbf{9.28}\\ 
		\bottomrule
\end{tabular}}
\caption{Results on the REVERIE dataset. MAttNet~\cite{mattnet} is adopted for object grounding. The top two results are highlighted in red bold and blue italic fonts, respectively.}
\label{tab:reverie}
\end{table*}

\subsection{Implementation details}
We set parameters $\lambda_1$ and $\lambda_2$ to 0.2 to keep the associated losses at the same level with others, and set $\lambda_3$ to 0.2 following \cite{envdrop}. We use $U=12$ transformers in our sequential BERT module. To take advantage of existing BERT-based works, we initialise the transformers in our model with the weights of UNITER~\cite{uniter}. UNITER is originally trained on four image-text datasets (COCO~\cite{coco}, Visual Genome~\cite{vg}, Conceptual Captions~\cite{cc}, and SBU Captions~\cite{sbu}) for joint visual and textual understanding. Note that these datasets have no overlap with the VLN data.
Our model is trained separately for each task. We use the AdamW~\cite{adam} optimiser with a learning rate of $1\times 10^{-6}$ and the model is trained on 8 Nvidia V100 GPUs. For the R2R task, following common practice, augmentation data from Envdrop~\cite{envdrop} are used. For the REVERIE and NDH tasks, only the original training data are used. It is worth noticing that although Thomason \etal~\cite{ndh} show that using the complete dialogue leads to much better results on the NDH task (rather than using just the oracle's answer), here we choose the oracle's answer as instruction to avoid the additional computation cost, and we achieve the best results documented, as is demonstrated later.

\subsection{Evaluation Metrics}

\paragraph{Metrics for REVERIE.}  Trajectory Length (TL), Success Rate (SR), Success weighted by Path Length (SPL) and Oracle Success Rate (OSR) are commonly used metrics for the navigation sub-task. Navigation is considered successful if the target object can be observed at the agent's final location, and considered as oracle-successful if the target object can be observed at one of its passed locations.  TL is the average length of agent's navigation trajectories. SR is the percentage of successful tasks. SPL is the main metric for navigation~\cite{spl}, which represents a trade-off between SR and TL. Higher SPL indicates better navigation efficiency. % 
Additionally, REVERIE uses  Remote Grounding Success rate (RGS) and  {RGS} weighted by Path Length (RGSPL) as whole-task performance metrics to measure the percentage of tasks that correctly locate the target object. 

\vspace{-11pt}
\paragraph{Metrics for NDH.}  Goal Progress {(GP)} is the main metric~\cite{ndh}, which measures how much progress (in meters) the agent has made towards the target. A higher {GP} denotes a better performance.

\vspace{-11pt}
\paragraph{Metrics for R2R.} {TL}, {SR}, {SPL} and Navigation Error {(NE)} are four widely adopted metrics. A navigation is considered as successful if the agent stops within 3 meters to the target. {NE} is the average distance between the agent's final location and the target.

\subsection{Comparison to State-of-the-Art Methods}

% --------- 
\begin{table}[!t]
\centering
\resizebox{\linewidth}{!}{
	\begin{tabular}{lccc|ccc}
		\toprule
		\multirow{2}{*}{Agent} & \multicolumn{3}{c|}{Val Unseen} & \multicolumn{3}{c}{Test Unseen} \\ \cline{2-7} 
		& Oracle & Navigator & Mixed & Oracle & Navigator & Mixed \\ \hline
		Random & 1.09 & 1.09 & 1.09 & 0.83 & 0.83 & 0.83 \\
		Skyline & 8.36 & 7.99 & 9.58 & 8.06 & 8.48 & 9.76 \\
		Seq2Seq~\cite{r2r} & 1.23 & 1.98 & 2.10 & 1.25 & 2.11 & 2.35 \\
		CMN~\cite{cmn} & \bif{2.68} & \bif{2.28} & \bif{2.97} & \bif{2.69} & \bif{2.26} & \bif{2.95} \\ 
		\rowcolor{mygray} PREVALENT~\cite{prevalent}&2.58 &2.99 &3.15&1.67&2.39&2.44\\
		\hline
		ORIST & \rbf{3.30} & \rbf{3.29} & \rbf{3.55} & \rbf{2.78} & \rbf{3.17} & \rbf{3.15} \\ \bottomrule
\end{tabular}}%
\caption{Results of the proposed method on the NDH datasets compared against state-of-the-art methods. Note that PREVALENT is pre-trained on a large scale augmented VLN dataset while ours not. The top two results are highlighted in red bold and blue italic fonts, respectively.}
\label{tab:ndh}
\end{table}

\begin{table}[!t]
\centering
\resizebox{\linewidth}{!}{%
	\begin{tabular}{lcccc|cccc}
		\toprule
		\multirow{2}{*}{Agent}  & \multicolumn{4}{c|}{Val Unseen} & \multicolumn{4}{c}{Test} \\ \cline{2-9} 
		& TL$\downarrow$ & NE$\downarrow$ & SR$\uparrow$ & \textbf{SPL}$\uparrow$ & TL$\downarrow$ & NE$\downarrow$ & SR$\uparrow$ & \textbf{SPL}$\uparrow$ \\ \hline
		
		FAST~\cite{fast}  & 21.17 & 4.97 & 56 & 43 & 22.08 & 5.14 & 54 & 41 \\
		
		SM~\cite{selfmonitor} & 17.09 & - & 43 & 29 & 18.04 & 5.67 & 48 & 35 \\
		
		EnvDrop~\cite{envdrop} & 10.70 & 5.22 & 52 & 48 & 11.66 & 5.23 & 51 & 47 \\
		
		AuxRN~\cite{auxiliary} & - & 5.28 & 55 & \bif{50} & - & 5.15 & 55 & \bif{51} \\
		
		OAAM~\cite{oaam} & 9.95 & - & 54 & \bif{50} & 10.40 & 5.30 & 53 & 50 \\
		
		SERL~\cite{softexp} & - & 4.74 & 56 & 48 & 12.13 & 5.63 & 53 & 49 \\
		
		AVIG~\cite{avig} & 20.60 & 4.36 & 58 & 40 & 21.60 & 4.33 & 60 & 41 \\
		\rowcolor{mygray} PREVALENT\cite{prevalent} &  10.19 & 4.71 & 58 & 53 & 10.51 & 5.30 & 54 & 51 \\
		\hline
		ORIST & 10.90 & 4.72 & {57} & \rbf{51}  & 11.31 & 5.10 & 57 & \rbf{52} \\ 
		\bottomrule
	\end{tabular}%
}
\caption{Single run results on the R2R dataset compared against state-of-the-art methods. 
	The top two results are highlighted in red bold and blue italic fonts, respectively.}
\label{tab:r2r}
\end{table}

\paragraph{Results on REVERIE.}
REVERIE is a newly proposed VLN task~\cite{reverie}.  
We evaluate our method against three baselines and a state-of-the-art model provided in~\cite{reverie}. Our method first performs navigation and then conducts object grounding as in~\cite{reverie} after the agent stops.

Table~\ref{tab:reverie} presents the evaluation results. 
The proposed ORIST method achieves the best performance on both the Val UnSeen and Test splits. Particularly, our ORIST obtains nearly double the previous best performance in terms of the main SPL and RGSPL metrics on the Val UnSeen split. On the test split, it also achieves approximately 7$\%$ and 3$\%$ absolute improvement in terms of the navigation metric SPL and the remote object grounding metric RGSPL, respectively. As analysed later in the ablation study, this mainly arises from the sequential design for the base BERT model and the direction loss. We also note that its navigation performance on the Val Seen split ranks second, which indicates our ORIST still has room to further learn from the training data.

\paragraph{Results on NDH.} NDH is a recently proposed VLN task. We compare against CMN~\cite{cmn} and PREVELANT~\cite{prevalent}. % 
We also compare against Seq2Seq~\cite{r2r}, the baseline proposed along with the NDH dataset.

%----- 
\begin{table*}[!t]
\centering
\resizebox{\linewidth}{!}{
	\begin{tabular}{ccccc|c|c|cccc|cccc}
		\toprule
		\multicolumn{5}{c|}{{Module}} & \multicolumn{1}{c|}{{NDH Val Seen}} &\multicolumn{1}{c|}{{NDH Val Unseen}} & \multicolumn{4}{c|}{{REVERIE Val Seen}} & \multicolumn{4}{c}{{REVERIE Val Unseen}}\\ 
		\hline
		&S & $\mathcal{L}^\mathbf{D}$ & $\mathcal{L}^{\mathbf{R}_n}$ &$\mathcal{L}^{\mathbf{R}_g}$ & \textbf{GP}$\uparrow$ &  \textbf{GP}$\uparrow$ & SR$\uparrow$ & OSR$\uparrow$ & \textbf{SPL}$\uparrow$ & TL$\downarrow$ &SR$\uparrow$ & OSR$\uparrow$ & \textbf{SPL}$\uparrow$ & TL$\downarrow$\\ \hline
		\#1&  & &  &  &4.00& 2.42& 41.9&47.1& 39.5&10.83&9.50&23.9&7.7&12.43\\
		\#2 &\cmark  &  &  &  & 3.80 &2.86  & 41.4 & 44.9 &40.1 &10.95&15.1&21.7&13.3&11.23\\
		\#3 &\cmark  &\cmark  & & & 3.88 &3.09  & 41.5 & 44.3 &39.5 &10.64&18.6&24.0&16.9&10.41 \\
		\#4 &\cmark  &\cmark &\cmark & & 4.12 & 3.15 &42.9& 46.9& 40.3&10.96&17.1&24.3&14.9&12.24 \\
		\#5 &\cmark  &\cmark &\cmark &\cmark &4.21& 3.30&45.2&49.1& 42.2&10.73&16.8&25.0&15.1&10.90 \\
		\bottomrule
	\end{tabular}
}
\caption{Ablation study of our model. Symbols S, $ \mathcal{L}^{\mathbf{D}}, \mathcal{L}^{\mathbf{R}_n},\mathcal{L}^{\mathbf{R}_g}$ denotes our sequential design, direction loss, room-type loss for the next step, and room-type loss for the goal location, respectively.}
\label{tab:ablation}
\end{table*}

Table~\ref{tab:ndh} presents the navigation results using GP as the metric in three settings. The results show that our ORIST method achieves the best performance on both Val Unseen and Test Unseen splits under all settings. Specifically, ORIST achieves about $0.9$ meters absolute improvement (about 40$\%$ relative improvement) under the ``Navigator'' setting on the Test Unseen split. 
As shown later in the ablation study, both our sequential design and new losses contribute to the final success.

\vspace{-8pt}
\paragraph{Results on R2R.} We compare our model against eight state-of-the-art navigation methods, including FAST~\cite{fast}, SelfMonitor~\cite{selfmonitor}, EnvDrop~\cite{envdrop}, AuxRN~\cite{auxiliary}, OAAM~\cite{oaam}, SERL~\cite{softexp}, AVIG~\cite{avig}, and PREVALENT~\cite{prevalent}. PREVALENT~\cite{prevalent} is pre-trained on a large scale augmented VLN dataset,
and hence is not directly equivalent.   VLN-BERT~\cite{bertvln} is not compared because it needs to obtain a set of candidate paths via extra VLN methods.
Table~\ref{tab:r2r} presents the comparison of navigation results. 
Our ORIST method achieves favorable results on the Test split in terms of the main metric SPL, even taking PREVALENT into consideration.

\subsection{Ablation Study}
In this section, we evaluate the impact of the key components of our model: the sequential design for BERT and the newly proposed three loss functions. The evaluations are conducted on the NDH and REVERIE datasets. NDH utilises detailed navigation instructions as in the R2R task, while REVERIE utilises much shorter and high level instructions.
The results are shown in Table~\ref{tab:ablation}, where the symbols S, $\mathcal{L}^{\mathbf{D}}, \mathcal{L}^{\mathbf{R}_n}$, and $\mathcal{L}^{\mathbf{R}_g}$ denote the sequential design for BERT,  direction loss, room-type loss for the next step, and room-type loss for the goal location, respectively.

\vspace{-12pt}
\paragraph{Effectiveness of the Sequential Design.} To evaluate the effectiveness of the proposed sequential design, we compare against its variant which is a direct application of BERT (\ie, without the connection from step $t$ to step $t+1$). Both are initialised from UNITER.
The results are presented in rows $1$ and $2$ of Table~\ref{tab:ablation}. They show that: (I) For Val Unseen splits, our sequential design achieves significant improvements:  $0.44$ meters absolute improvement ($18\%$ relative improvement) for NDH task and $6\%$ absolute improvement ($73\%$ relative improvement) for REVERIE task. (II) For Val Seen splits, a slight performance regression is observed for the NDH task and a slight improvement appears on the REVERIE task. These results show that our sequential design effectively enhances the generalization ability and may reduce over-fitting in seen environments.

\vspace{-8pt}
\paragraph{Effectiveness of Direction and Room-type Losses.} By comparing the results in rows $2$ and $3$ of Table~\ref{tab:ablation}, we observe that the direction loss $\mathcal{L}^{\mathbf{D}}$ brings consistent performance improvement on all four splits (in terms of SR for REVERIE), and particularly on the unseen splits. % 
This might be attributed to that the direction information increases the distinguishability of navigation candidates and is easily generalized to unseen scenarios.
Regarding the room-type losses (rows 3, 4, and 5), we find that for the NDH task both losses boost the performance (\eg, from 3.88 to 4.21 on Val Seen, and from 3.09 to 3.3 on Val Unseen); for the REVERIE task, both losses improve performance on the Val Seen split (the SPL score increases from 39.5 to 42.2, SR from 41.5 to 45.2) while the next-room type loss $\mathcal{L}^{\mathbf{R}_n}$ harms on Val Unseen split. This might be caused by the fact that there is no next room mentioned in REVERIE. In contrast, its improvement on the Val Seen split could arise from over-fitting to some seen scenarios.

\begin{figure*}[!t]
\centering
\includegraphics[width=\linewidth]{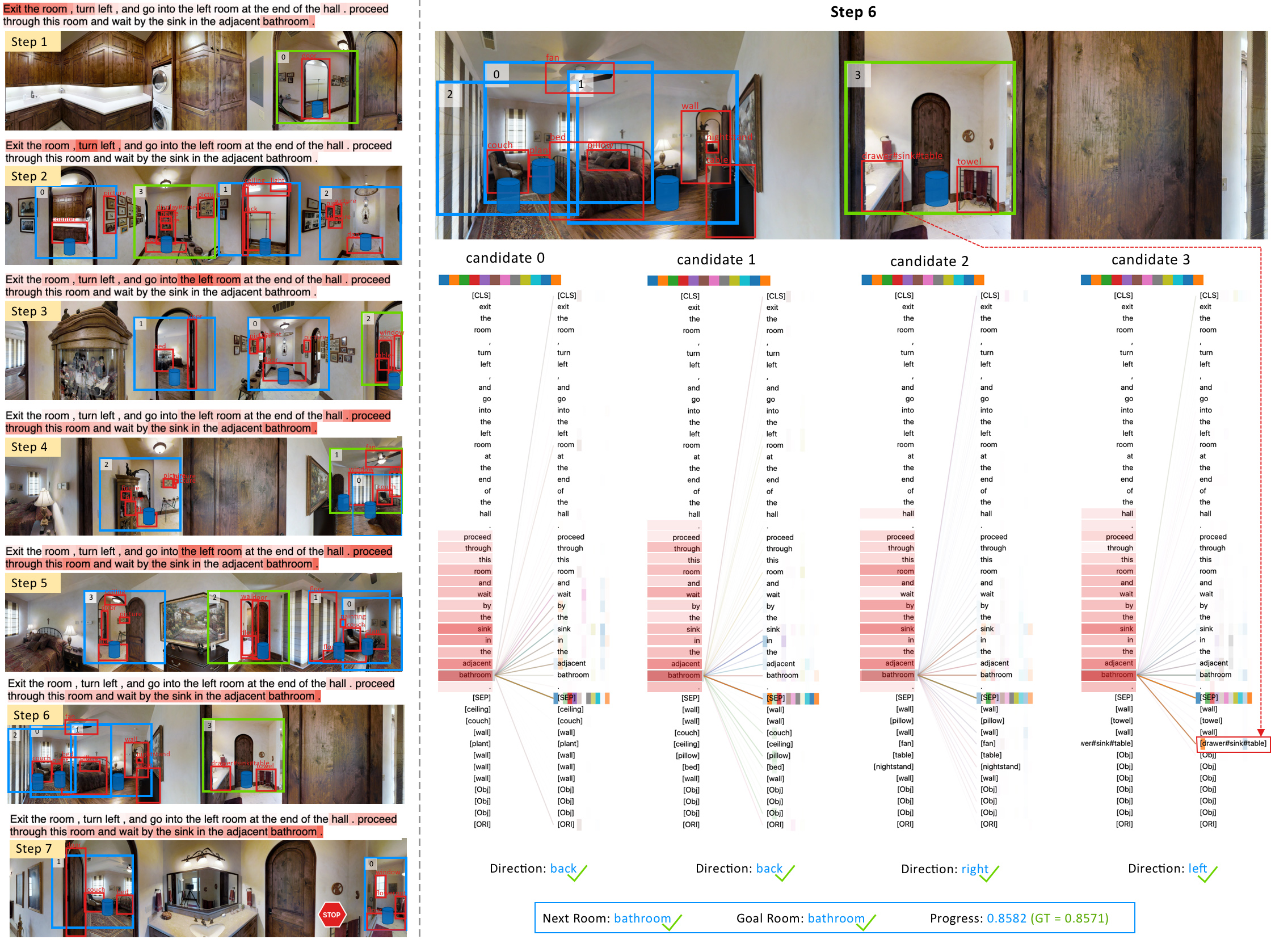}
\caption{Visualisation of a navigation trajectory. The left panel depicts the whole trajectory, and the right panel shows attention distributions of a specific step (see Section~\ref{sec:qua_res} for details). Object tokens are indicated by ``[ ]'', and  ``[ORI]'' denotes the orientation token.
}
\label{fig:visulisation}
\end{figure*}

\subsection{Qualitative Results}
\label{sec:qua_res}
In Figure~\ref{fig:visulisation}, we visualise one navigation trajectory on the R2R task to give an intuitive example of how our agent navigates. The visualisation is based on the attention visualisation tool BertViz~\cite{bertviz}. 
The left panel shows the process of attention shifting as the agent navigates. Navigable locations at each step are marked using blue cylinders and the perceived views for each navigable location are bounded using an indexed rectangle. The green rectangle denotes the view associated with the next step predicted by our model. Objects in each view are marked with red bounding boxes. 
As shown in the left panel, the attention on the instruction shifts intuitively as the agent navigates. Concretely, at the first step, the agent focuses on ``exit the room'' and it goes forward to the door. Next, the agent becomes aware of ``turn left'' and performs the correct action. Then, at steps 3$\sim$5, the attention is updated from ``the left room'' to ``proceed through the room'', and the agent behaves accordingly. 
At the last two steps, the attention focuses on the end of the instruction and the agent recognises the bathroom by virtue of the presence of the sink. 
Hence, it chooses the correct viewpoint and finally decides to stop. 

To illustrate that the agent is on the road to ``know where'', we take the decision progress of step 6 as an example and show  details in the right panel.
Specifically, 
the right panel shows two kinds of attention distributions under each navigation candidate: 1) Attention on the instruction. As shown by the pink color on the left column words (the darker the larger), the ``bathroom'' is focused on. 2) Self-attention on all tokens. In particular, the right column shows attention distribution of  ``bathroom'' on all tokens. We can see ``bathroom'' is strongly connected to the object ``[drawer\#sink\#table]'' (a darker line between two tokens denotes a stronger connection). The 12 kinds of colors denote attention computed from 12 attention heads.
Additionally, at the bottom of the right panel, we  present the progress prediction, the next room  and goal room type prediction, and the direction of each navigable viewpoint.  The agent successfully infers these clues from the instruction and visual perceptions.  All of these information assist the agent making the correct navigation decision.

\section{Conclusion}
We have proposed a novel unified sequential BERT model for general indoor VLN tasks. The sequential characteristic enables the BERT to be better applied to several VLN tasks.
By taking object-level and word-level inputs, our model is able to learn fine-grained relationships across textual and visual modalities for VLN.
Moreover, we design a new direction loss and two room-type losses to facilitate the model to predict more accurate navigation actions. 
Extensive experimental evaluations on three VLN tasks demonstrate the effectiveness and generalisation ability of the proposed method. 

As our sequential BERT is able to directly predict navigation actions,
it enables us to combine the multi-task VLN pre-training and downstream task fine-tuning.
We leave this as a future work.

\section{Acknowledgements}
This work is supported in part by the ARC DE190100539 and the NSF CAREER Grant \#1149783.

\clearpage
{\small
\bibliographystyle{ieee_fullname}
\bibliography{egbib}
}

\end{document}